\documentclass[a4paper]{article}
\usepackage{Odyssey2024}
\usepackage{epsfig,amssymb,amsmath}
\ninept
\usepackage{amsmath,graphicx,cuted,capt-of}

\usepackage{booktabs}
\usepackage{url,float,afterpage}
\usepackage{placeins}  
\usepackage{tcolorbox}
\usepackage{enumitem}
\setlist{leftmargin=1mm}
\usepackage{pifont}
\usepackage{amsmath}
\usepackage{setspace}

\title{Improved Contextual Recognition in Automatic Speech Recognition Systems by Semantic Lattice Rescoring}

\name{Ankitha Sudarshan$^1$, Vinay Samuel$^2$, Parth Patwa$^3$, Ibtihel Amara$^4$, Aman Chadha$^{5,6\dag}$}

\address{$^1$Purdue University $^2$Carnegie Mellon University $^3$University of California Los Angeles\\ 
  $^4$McGill University $^5$Stanford University $^6$Amazon AI\\
{\small \tt $^1$sudarsh0@purdue.edu \tt $^{5,6}$hi@aman.ai} }
\begin{document}
\maketitle
\renewcommand{\thefootnote}{\fnsymbol{footnote}}
\footnotetext[2]{Work does not relate to position at Amazon.}
\renewcommand*{\thefootnote}{\arabic{footnote}}

\begin{abstract}
Automatic Speech Recognition (ASR) has witnessed a profound research interest. Recent breakthroughs have given ASR systems different prospects such as faithfully transcribing spoken language, which is a pivotal advancement in building conversational agents. However, there is still an imminent challenge of accurately discerning context-dependent words and phrases. 
In this work, we propose a novel approach for enhancing contextual recognition within ASR systems via semantic lattice processing leveraging the power of deep learning models in accurately delivering spot-on transcriptions across a wide variety of vocabularies and speaking styles. 
Our solution consists of using Gaussian Mixture Models and Hidden Markov Models (GMM-HMM) along with Deep Neural Networks (DNN) models integrating both language and acoustic modeling for better accuracy. We infused our network with the use of a transformer-based model to properly rescore the word lattice achieving remarkable capabilities with a palpable reduction in Word Error Rate (WER). We demonstrate the effectiveness of our proposed framework on the LibriSpeech dataset with empirical analyses.

\end{abstract}

\section{Introduction}
Recognizing spoken language accurately and efficiently is a complex task due to variability in the source of speech such as pronunciation, dialects, vocabulary, accents, articulation, etc. This variability presents significant challenges in the development and fine-tuning of Automatic Speech Recognition (ASR) systems. Different speakers can produce vastly different speech signals even when uttering the same word or phrase, complicating the task of accurate speech recognition. Factors such as background noise, speaker distance from the microphone, and the acoustic environment further exacerbate these challenges, necessitating sophisticated solutions to ensure ASR systems can perform reliably under a wide range of conditions.

The importance of addressing these variabilities cannot be overstated, as they significantly affect the accuracy and reliability of ASR systems. These systems are increasingly becoming integral in various applications, from virtual assistants to real-time translation services, thereby amplifying the need for improved precision. Moreover, the rising demand for hands-free operations and accessibility features for individuals with disabilities further underscores the urgency of refining ASR technology. To this end, researchers and developers are continuously exploring innovative approaches to enhance the robustness and adaptability of ASR systems, aiming to minimize errors and misunderstandings that can arise from speech variability.

Furthermore, the advent of global digitalization has led to an unprecedented intermingling of cultures and languages, introducing additional layers of complexity to speech recognition. The diversity in accents and dialects, even within the same language, poses a considerable challenge for ASR systems. This diversity necessitates the development of sophisticated models that can adapt to the nuances of spoken language across different regions and communities.

Additionally, the rapid expansion of smart home devices relies heavily on effective speech recognition capabilities. As these technologies become more embedded in our daily lives, the ability of ASR systems to accurately interpret commands and queries becomes crucial for user experience. Incorrect recognition can lead to frustration, errors, and in some contexts, serious consequences.

The interaction between humans and machines through voice commands is predicated on the ability of the system to understand the intent and context of the spoken words. This interaction becomes even more challenging in noisy environments or when the speech signal is distorted, highlighting the need for robust ASR systems that can perform well under less-than-ideal conditions.

Technological advancements in artificial intelligence have opened new frontiers in enhancing ASR accuracy. However, the pace at which spoken language evolves, with new slang, terminologies, and phrases emerging constantly, presents an ongoing challenge. Keeping ASR systems updated and relevant requires not just technological innovation but also a deep understanding of linguistic dynamics and cultural nuances.

Semantic interpretation of a speech is crucial in ASR systems. Let us take the following example. \emph{``I am going to a bank to deposit a check"}. Without context, the word bank could refer to either a financial institution or either the edge of a river illustrating the critical need for ASR systems to contextualize language effectively. This ambiguity underscores the complexity of human language and highlights the limitations of current ASR systems in understanding context and disambiguating meaning. By addressing this gap, we can significantly improve the functionality and applicability of ASR technology across a broad spectrum of use-cases. To bridge this contextual gap in ASR systems, semantic lattice processing is a key component in contributing to better recognition of situational context conditions. This technique utilizes a lattice structure to represent the relationships between words and phrases in a sentence. It is created by analyzing the audio input and identifying possible word and phrase combinations and their associated probabilities. This information is then used to create a graph-like structure, where each node represents a word or phrase, and the edges represent the relationships between them \cite{deep3}. 

In this study, our primary emphasis centers on lattice re-scoring, a technique designed to efficiently re-evaluate the likelihood of potential speech hypotheses. While numerous lattice re-scoring techniques have been documented in the literature \cite{semlat,kumar2017lattice}, our research introduces a novel approach tailored to bolster contextual information within ASR systems.

Our key contributions can be summarized as follows:

\begin{tcolorbox}[colback=blue!5!white,colframe=blue!75!black,title={\textbf{\textsc{{Our Contributions}}}}]
\begin{itemize}
\item[\ding{224}]  {\footnotesize 
    {\rmfamily\fontsize{8}{9}\selectfont
Lattice re-scoring, which refines recognition results through the integration of language model probabilities from different language models, enhancing transcription accuracy and overall system performance.
}}
\vspace{2mm}
\item[\ding{224}]  {\footnotesize 
     {\rmfamily\fontsize{8}{9}\selectfont
Employing a Transformer architecture for our neural language model, enhancing lattice re-scoring with state-of-the-art contextual modeling.
}}
\vspace{2mm}
\item[\ding{224}]  {\footnotesize 
     {\rmfamily\fontsize{8}{9}\selectfont
Achieving a 1.36\% reduction in word error rate when compared to state-of-the-art models sharing a similar architectural framework.
}}
\end{itemize}
\end{tcolorbox}


\section{Related Work}
\label{sec:format}

Voice assistants use a variety of end-to-end (E2E) ASR techniques, including attention-based encoder-decoder (AED) \cite{ chan2016listen, dong2018speech}, recurrent neural network transducer (RNN-T) \cite{graves2006connectionist}, and connectionist temporal classification (CTC) \cite{graves2006connectionist}. During training, the E2E model simultaneously optimizes the whole recognition pipeline and produces the word sequence output directly.  
One problem with this method, though, is that it has trouble identifying terms like songs or human names that don't often appear in the training set.

ASR systems' contextual recognition is crucial, especially for voice assistants, which must identify names of contacts, musicians in a user's music collection, and other entities. Shallow fusion \cite{shallow1, shallow5}, attention-based deep context \cite{deep1,deep4,deep3,Wang_2020}, and trie-based deep biasing \cite{trie1, trie2} are the current contextual biasing techniques used for various E2E \cite{L_scher_2019} ASR models. 

According to \cite{deep3, deep4}, phoneme information may be injected or training with challenging negative examples can help to decrease misunderstanding between similar words. On-the-fly re-scoring is the most popular method for introducing contextual bias into ASR. In \cite{hall2015composition}, this technique was first used with hybrid ASR models. In order to allow the weights on the bias terms to be changed ``on the fly" at the time of inference, it entails composing a Weighted Finite State Transducer (WFST) that represents the ASR model with a novel WFST representation of the bias terms. The bias terms are still assembled into a WFST representation for E2E models, but the ASR model composes that WFST at each decoding time-step, determining the likelihood of the current hypothesis based on the ASR model is combined with a score from the bias WFST.

A separate re-scoring module can not adequately handle the error introduced by the upstream acoustic network.  Some architectures \cite{deep1, chang2021context} can integrate contextual information into acoustic model to improve the output posterior distribution corresponding to the related contextual words and fit the subsequent external LM well. As the size of the possible contextual word list grows in real applications, the accuracy and latency of the system descend rapidly due to dispersed attention score and heavy attention computation. Moreover, in practice, it is hard to obtain compact and accurate contextual information in advance. Consider music search - we may face a large contextual word list (size in thousands) containing popular songs. In this case, E2E context bias cannot work well due to the large size and low quality of the contextual word list, leading to performance degradation \cite{han2022improving}.

The contextual biasing\cite{fu2023robust,Huang2020ClassLA,shenoy2021contextual} method often struggles with accurately processing and recognizing OOV (Out Of Vocabulary) \cite{parada-etal-2010-contextual} terms, which are prevalent in real-world scenarios, especially in applications like voice-controlled music systems. Our approach introduces semantic lattice rescoring using Kaldi, a technique designed to refine predictions and improve the overall accuracy of the ASR system. By systematically addressing these limitations, our paper offers a significant contribution to the field, showcasing an innovative strategy to enhance contextual recognition, particularly when dealing with challenging OOV terms like song titles.

\begin{figure}[ht]
    \centering
    \includegraphics[width=\linewidth]{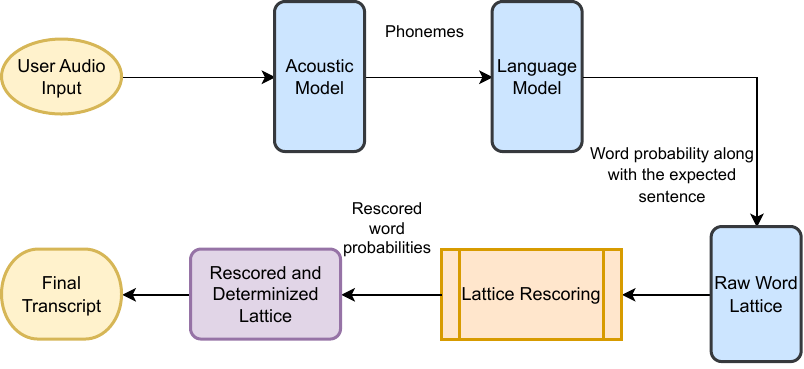}
    \caption{\textbf{Global overview of our framework.} Our framework includes audio input, DNN acoustic model,language model integration, lattice creation and alignment, transformer re-scoring, and transcript generation.}
    \label{fig:framework_final}
\end{figure}

\begin{figure}[ht]
    \centering
    \includegraphics[scale=0.55]{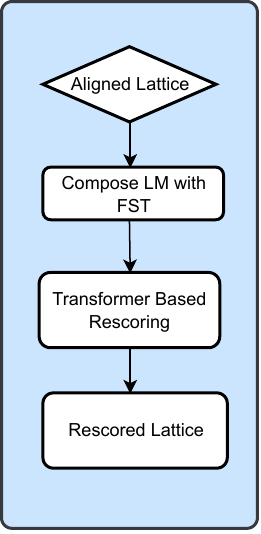}   
    \caption{\textbf{Lattice re-scoring strategy.} Alignment is involved, n-gram language model integration, and Transformer-based re-scoring to enhance contextual features in the final transcript.}
    \label{fig:lattice}
\end{figure}

\section{Methodology}
\label{sec:methodology}
\subsection{Background}

In the realm of Automatic Speech Recognition (ASR) systems, the initial decoding phase marks the inception of a lattice—a directed acyclic graph (DAG) that intricately captures a spectrum of potential word hypotheses, each accompanied by its respective score. This lattice serves as a comprehensive representation, embodying the rich diversity of candidate hypotheses in the form of a structured graph. However, the inherent complexity of spoken language often results in the lattice containing a multitude of potential paths, necessitating further refinement to enhance the accuracy of ASR outcomes.

Lattice re-scoring emerges as a critical post-processing mechanism designed to elevate the precision of ASR hypotheses within the lattice structure. By re-evaluating and re-ranking potential word sequences, this process aims to discern the most plausible and contextually fitting interpretations of the spoken input. Our investigative efforts delve into the intricate mathematical and algorithmic dimensions of lattice re-scoring, seeking to unravel the underlying principles that govern this transformative phase of ASR.

To undertake this exploration, we deploy a Transformer model, meticulously configured with positional encoding to capture temporal relationships. The model features multiple Transformer encoder layers, parameterized by $(n$), an input embedding layer, and an output linear layer dedicated to sequence prediction. This tailored architecture stands as the bedrock of our investigation, allowing us to probe the synergistic interplay between lattice re-scoring and the transformative capabilities inherent in the Transformer model.

Through a comprehensive approach, we aim to dissect the intricacies of lattice re-scoring, shedding light on its multifaceted role in refining the accuracy and efficacy of ASR systems.

Let $A(a)$ represent the acoustic score for lattice arc $a$, indicating the alignment of audio with its associated word. The language model probability $P(w_1, w_2, ..., w_N)$, derived from a Transformer model, signifies the likelihood of the entire word sequence $(w_1, w_2, ..., w_N)$ within the language context.
For a path $P$ through the lattice, the word sequence probability $P(P)$ is calculated as the product of acoustic scores and language model probabilities along the path: $P(P) = \prod_{a}(A(a) \cdot P(w))$.
Here, $Wa$ represents the word associated with arc $a$, introducing flexibility to address concerns regarding the fixed nature of ``w" in all arcs.
The optimal path $P*$ maximizing the joint probability of the word sequence is given by: $P* = argmax_P P(P)$.

\subsection{Lattice Re-scoring}

We provide in Figure \ref{fig:framework_final} and Figure \ref{fig:lattice}, respectively, our overall framework and the proposed lattice re-scoring strategy. We use the Deep Neural Networks (DNN)-refined predictions for creating word lattices as an intermediate representation of the ASR output. This decoding process generates lattice-like outputs that contain phone-level alignments and associated scores and eventual word alignments.

Each path in the lattice has a score based on the ASR system's confidence. However, we can improve the transcription by reevaluating these paths using a neural LM\cite{6854535}, which captures the likelihood of different word sequences based on linguistic context. We use our custom-trained transformer to perform re-scoring. A Transformer-based re-scoring \cite{yang2022conformer,pandey2021lattention} approach, as opposed to traditional n-gram methods, introduces novelty by leveraging advanced neural network architectures that are designed to handle sequences more effectively and capture complex language patterns.

This transformation is done by computing the conditional probability of the word sequence in each path given the language model and combining it with the original acoustic likelihood score. The result is a new lattice where paths have modified scores that reflect both acoustic and language model information, enhancing transcription accuracy.

The scores from the neural LM are converted to log-likelihoods to be combined with the original lattice scores. Once we have the lattice paths re-scored using the neural LM and have converted the scores to log-likelihoods, we combine these scores with the original lattice scores. This step helps integrate the language model probabilities into the lattice.

By combining these scores, the lattice paths that were previously assigned lower scores by the ASR system but have higher probabilities according to the LM are promoted, resulting in a better transcription of the input audio than the original input. 

A prominent mathematical formula in the lattice creation process in ASR is related to the computation of the overall likelihood or score of a path through the lattice, defined in Equation \eqref{eq1}. This score is typically calculated as the sum of individual acoustic and language model scores along the path. 

\begin{equation}
\begin{aligned}
\text{Path Score} = \sum_{i=1}^{N} \Bigl(&\log\bigl(P(\text{word}_i | \text{word}_{i-1})\bigr) \\
&+ \log\bigl(P(\text{Acoustic features}_i | \text{word}_i)\bigr)\Bigr)
\label{eq1}
\end{aligned}
\end{equation}
where, $N$ represents the number of words in the path, $\text{word}_i$ is the $i^{th}$ word in the path, $P(\text{word}_i | \text{word}_{i-1})$ is the conditional probability of transitioning from $\text{word}_{i-1}$ to $\text{word}_i$ using the language model, and $P(\text{acoustic features}_i | \text{word}_i)$ is the probability of observing acoustic features at position $i$ given $\text{word}_i$ using the acoustic model.

\section{Experimental Details}
\subsection{Data Corpus and Preprocessing}
Our research leverages the LibriSpeech dataset \cite{librispeech}, a rich repository comprising around 1000 hours of meticulously curated English speech recordings, all sampled at a rate of 16 kHz. The dataset's substantial size and high-quality content serve as a foundational bedrock for our proposed methodology, ensuring not only the robustness of our approach but also its generalizability across diverse speech patterns and contexts.

To preprocess the data, we employ the versatile Kaldi toolkit. This step involves meticulously organizing the data into training, validation, and test sets, a crucial preparatory phase for subsequent model training and evaluation. The Kaldi format, adopted for its efficiency, comprises two integral components: the archive file (\texttt{.ark}) and the corresponding index file (\texttt{.scp}). The archive file encapsulates binary data, organized in a sequence of key-value pairs. Here, the key serves as a unique identifier, typically a string associated with the data segment, while the value represents the binary data itself.

Our data preparation encompasses the creation of text transcriptions and corresponding acoustic features for each segment of the audio data. This meticulous process not only establishes the groundwork for training our models but also lays the foundation for a comprehensive understanding of the linguistic and acoustic intricacies embedded within the dataset.

\subsection{Acoustic Model}

In the initial stages of our research, we begin with the preprocessing phase as the foundational step in our methodology. Building upon this foundation, our approach seamlessly integrates the sophisticated Gaussian Mixture Models and Hidden Markov Models (GMM-HMM) acoustic framework. The GMM-HMM model, a cornerstone of our methodology, undergoes a meticulous training process leveraging the acoustic features extracted during the preprocessing stage.

A strength of the GMM-HMM model lies in its inherent capacity to yield posterior probabilities associated with subword units, such as phonemes or context-dependent acoustic states. This capability extends to providing detailed probabilistic information for each individual frame of audio within the dataset. The richness of this probabilistic representation offers a granular view of the intricate acoustic characteristics embedded in the audio data, contributing significantly to the model's ability to capture nuanced patterns.

The generation of these posterior probabilities leads to the formulation of a meticulously constructed likelihood matrix. Within this matrix, the probabilities are systematically organized, providing an intricate overview of the distribution of acoustic features across various subword units throughout the entire audio sequence. This likelihood matrix serves as a valuable resource, offering not only a detailed snapshot of the acoustic landscape but also a foundation for further analyses.

The nuanced information encapsulated in the likelihood matrix is integral to our research, furnishing a thorough understanding of the temporal evolution of acoustic characteristics within the framework of GMM-HMM. By scrutinizing this matrix, we gain insights into how subword units evolve over time, providing a comprehensive perspective on the dynamics of acoustic features throughout the duration of the audio dataset.

\begin{strip}
   
      \includegraphics[width=\textwidth]{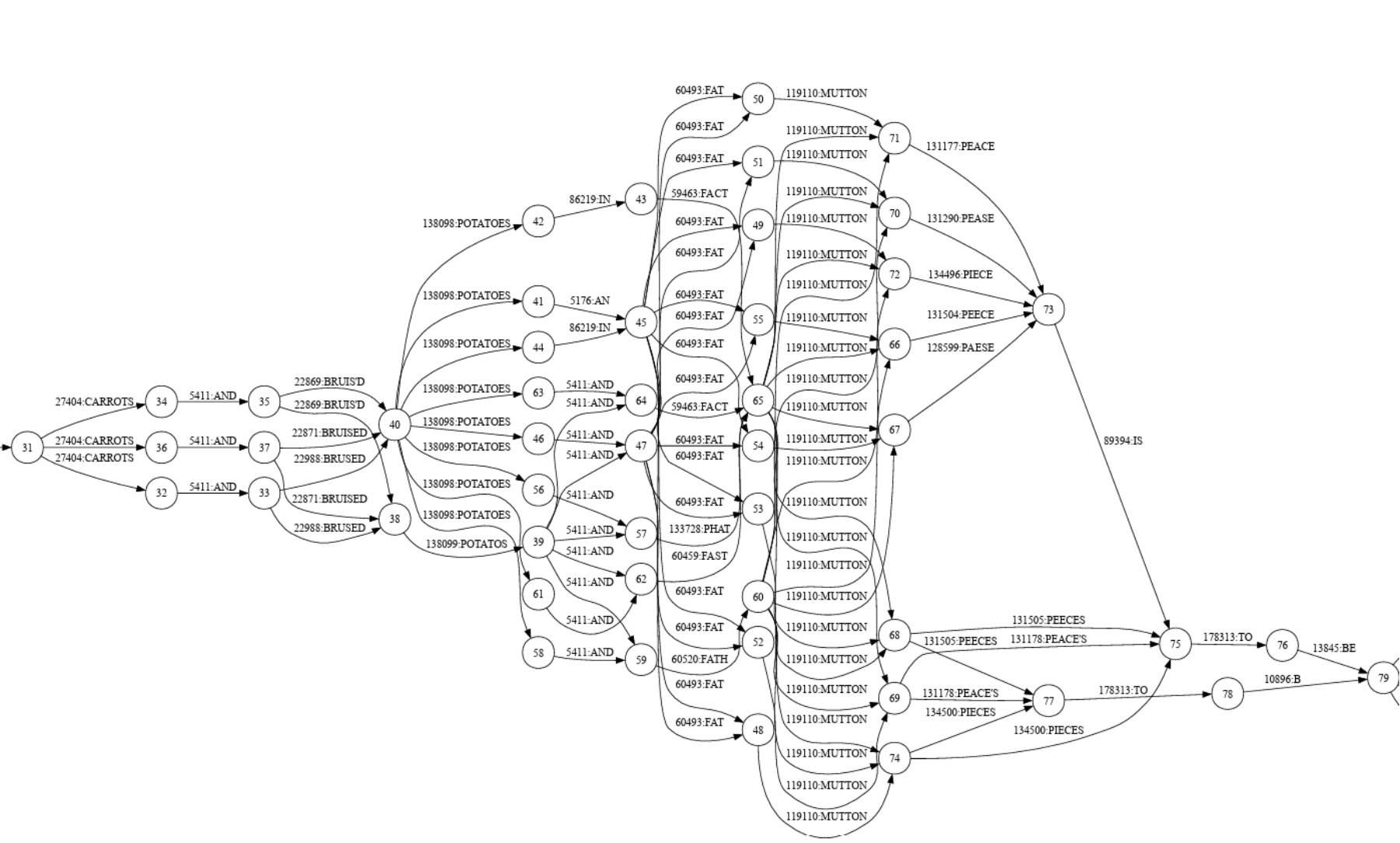}
      \captionof{figure}{A generic view of the lattice}
      \label{fig:magnifiedattice2}

\end{strip}

This probabilistic mapping augments the model's capability to discern subtle nuances within the audio data, culminating in a more refined and accurate representation of the underlying acoustic structures. This methodological approach underscores the meticulous attention given to extracting meaningful information from the acoustic domain, contributing to the robustness and efficacy of our overall framework.
\subsection{Language Model: Deep Neural Network (DNN)}
Following the GMM-HMM stage, we incorporate a DNN  to refine and improve the predictions made by the GMM-HMM model.

The DNN model generated enhances posterior probabilities over subword units. This DNN-refined output provides more accurate representations of the spoken audio, building upon the GMM-HMM predictions. For instance, if the utterance ``The quick brown fox" has initial ASR scores as 0.8, 0.7, 0.6 and post DNN rescoring scores as 0.9, 0.8, 0.7, the final transcript would have a confidence score of 0.9.

We use a neural LM -- a custom transformer trained on the same Librispeech dataset for the rescoring task. We then compose Finite State Transducers (FSTs) with the Language Model FST. The FST created using Kaldi's tools represents the language model, lexicon, and any other components of the speech recognition system. We then compile the Hidden Markov Model, Context, Lexicon, Grammar (HCLG) FST using the trained acoustic and language models, lexicon, and other necessary components.

These word-level alignments are then converted into Finite State Transducers (FSTs). These FSTs provide a structured representation that allow for efficient manipulation of word-level alignments. To this end, we generate four types of lattices:
\begin{description}
\item \textbf{Type 1:} DNN based lattice (with phone alignments followed by word alignments).  
\item \textbf{Type 2:} GMM based lattice (with phone alignments followed by word alignments). 
\item \textbf{Type 3:} DNN-based lattice (with direct word alignments).
\item \textbf{Type 4:} GMM-based lattice (with direct word alignments).
\end{description}.
Figure \ref{fig:magnifiedattice2} above represents a lattice structure of Type 1 for the utterance 'He Hoped There Would Be Stew Fort Dinner Turnips And Carrots And Bruised Potatoes'.

In summary, our methodology entails a two-stage process involving GMM-HMM and DNN models for the enhancement of spoken language processing in Automatic Speech Recognition (ASR) systems. In the initial GMM-HMM stage, raw spoken audio is subjected to detailed processing, producing foundational predictions that encompass acoustic features and subword structures. Subsequently, a DNN is deployed to refine these predictions, with a primary focus on improving posterior probabilities over subword units while leveraging insights from the GMM-HMM stage. The sequential integration ensures that the DNN capitalizes on the foundational knowledge established by the GMM-HMM model, resulting in a coherent and refined representation of spoken audio.

\begin{strip}
    
      \includegraphics[width=1\textwidth]{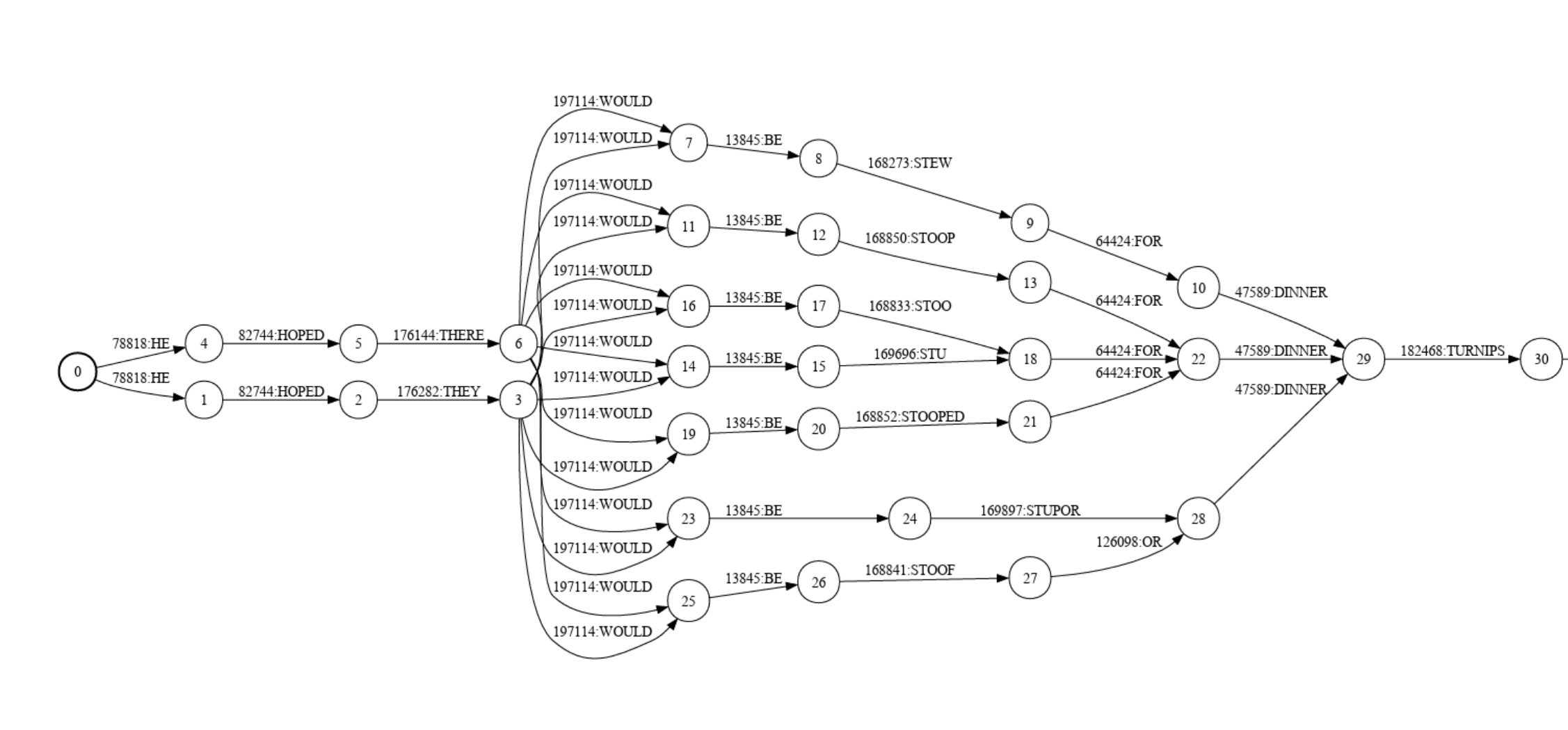}
      \captionof{figure}{Magnified view (partial) of the lattice }
      \label{fig:magnifiedattice1}

\end{strip}

The refined output from the DNN demonstrates a significant improvement over the GMM-HMM predictions, yielding more accurate representations of spoken language. This collaborative and sequential integration enhances the ASR system's capability to capture and interpret underlying patterns and structures in spoken language data, contributing to an overall improvement in the system's robustness and effectiveness.

\subsection{Transformer Model For Lattice Re-scoring}

In our experimental design, we tailored a six-layer transformer model, incorporating 512 hidden embeddings. This model underwent comprehensive training on an NVIDIA H100 GPU, leveraging the extensive LibriSpeech dataset as a foundational corpus for learning \cite{librispeech}. The training regimen spanned four epochs, characterized by a learning rate of 0.1 and a dropout rate of 0.1, ensuring a balanced yet robust learning process.

\begin{figure}[ht!]
  \includegraphics[width=1\linewidth]{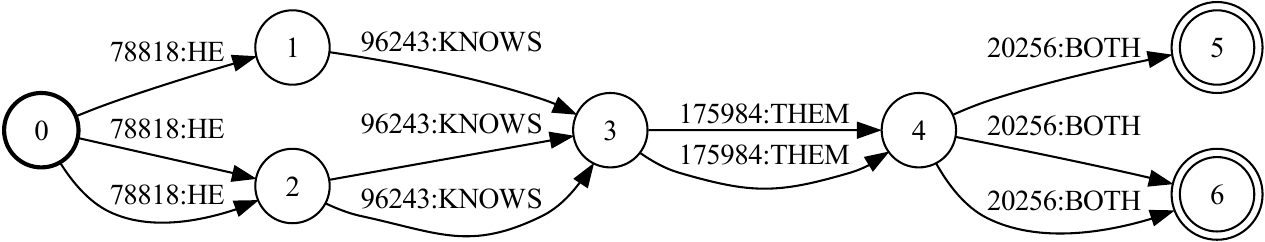}
  \caption{Example of pre-rescoring lattice given an input utterance.
  }
  \label{fig:speech_production1}
\end{figure}

\begin{figure}[ht!]
  \includegraphics[width=1\linewidth]{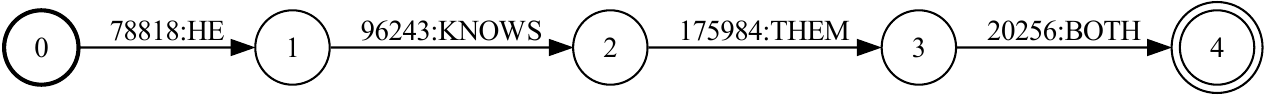}
  \caption{Example of post-rescoring lattice given an input utterance.}
  \label{fig:speech_production2}
\end{figure}

Visual insights into the transformative impact of our approach are encapsulated in Figures \ref{fig:speech_production1} and \ref{fig:speech_production2}, showcasing instances of lattices both pre- and post-transformative re-scoring. These visual representations indicate the evolution of lattices in response to the input utterance 'He knows them better.' A comparison between the pre-rescoring and post-rescoring lattices serves as a compelling visual testament to the significant enhancements achieved through our transformative approach.

The utilization of such illustrative examples not only substantiates the efficacy of our methodology but also provides a nuanced understanding of the intricate alterations and refinements introduced by the re-scoring process. 

\vspace{-3mm}
\section{Results and Discussion}
\label{sec:typestyle}

We conducted experiments on the four types of lattices mentioned in Section 4.3 and observed that their performance in the pipeline was identical. We thus decided to perform a comprehensive analysis based on Lattice Type 1 as a representative selection for the other lattice types. The below tables represent results for Lattice Type 1 on the test sets \texttt{`test-clean'} and \texttt{`test-other'}. 

\begin{table}[ht]
\centering
\begin{tabular}{ccccccc}
\toprule
LM Scale & \multicolumn{2}{c}{WIP 0} & \multicolumn{2}{c}{WIP 0.5} & \multicolumn{2}{c}{WIP 1} \\
 & (a) & (b) & (a) & (b) & (a) & (b) \\
\midrule
7 & 6.67 & 17.35 &6.65 &17.32 &6.67 &17.41 \\
10 & 6.75 &17.67 &6.76 &17.68 &6.79 &17.72 \\
13 &7.08 &18.35 &7.11 &18.39 &7.17 &18.43 \\
\bottomrule
\end{tabular}
\caption{WERs of Lattice Type 1 post-rescoring with three levels of word insertion penalties (WIPs) on both test datasets clean (a) and other (b).}
\label{tab:tab1}
\end{table}


\begin{table}[ht]
\centering
\begin{tabular}{ccccccc}
\toprule
LM Scale & \multicolumn{2}{c}{WIP 0} & \multicolumn{2}{c}{WIP 0.5} & \multicolumn{2}{c}{WIP 1} \\
 & (a) & (b) & (a) & (b) & (a) & (b) \\
\midrule

7 & 7.56 & 21.08 &7.64 &21.03 &7.67 &20.07 \\

10 & 8.16 &21.58 &8.21 &21.59 &8.31 &21.61 \\

13 &9.08 &23.04 &9.19 &23.12 &9.34 &23.16 \\
\bottomrule
\end{tabular}
\caption{WERs of Lattice Type 1 pre-rescoring (raw lattice) with three levels of word insertion penalties (WIPs) on both test datasets clean (a) and other (b).}
\label{tab:tab2}
    \vspace{-3mm}
\end{table}

From the results, we see a significant improvement in the Word Error Rate (WER) post rescoring. Considering the LM Scale-WIP combination of 7 and 0.5 respectively, we see a decrease of 14.88\% in the WER. This implies that our system performs slightly better than the SoTA of the HMM-(SAT) GMM model which has a WER of 8.01\% on the \texttt{test-clean} dataset and also beats the WER of 22.49\% on the \texttt{test-other} dataset.

In our results and discussion, it is evident that our rescoring architecture has played a pivotal role in achieving improved performance. Our rescorer, based on the Transformer architecture, proved to be a crucial component in improving WER. By leveraging the Transformer's attention mechanism, our model effectively captured long-range dependencies and contextual information within the input data. This enabled more accurate and contextually relevant predictions during the rescoring process, resulting in a notable reduction in WER. The model's ability to consider broader linguistic context played a pivotal role in achieving these enhanced ASR results.

\begin{table}[ht]
  \centering
          \resizebox{\columnwidth}{!}{
  \begin{tabular}{c|c|c|c|c}
    \toprule
    \multicolumn{5}{c}{LM Scale = 7} \\
    \multicolumn{5}{c}{WIP = 0.5} \\
    \midrule
    Test Set & Utterances (\#) & WER (\%) & Rescored WER (\%) & Change (\%) \\
    \midrule
    \texttt{test-clean} & 2620 & 7.64 & 6.65 & \textbf{-14.88}\\
    \texttt{test-other} & 2939 &  21.03 & 17.32 & \textbf{-21.42} \\
    \bottomrule
  \end{tabular}
  }
  \caption{WER on LibriSpeech for Kaldi ASR baseline.}
  \label{tab:wer_results}
\end{table}


\begin{table}[h!]
\vspace{-2mm}
\centering
\begin{tabular}{@{}lc@{}}
\toprule
\multicolumn{2}{c}{Technique} \\
\cmidrule(r){1-2}
Models with similar architecture & WER \\
\midrule
AmNet \cite{macoskey2021amortized} & 8.60 \\
HMM-SAT-GMM \cite{kaldi} & 7.19 \\
Snips \cite{coucke2018snips} & 6.40 \\
\midrule
Models with different architecture & WER \\
\midrule
Deepspeech2 \cite{amodei2015deep} & 5.83 \\
CTC + policy learning \cite{zhou2017improving}& 5.42 \\
Li-GRU \cite{ravanelli2019pytorchkaldi} & 6.20 \\
\hline
Ours & 6.65 \\
\bottomrule
\end{tabular}
\caption{A comparative study between our framework and the current SoTA for different ASR Models.}
\label{tab:mytable}
\end{table}

While our ASR model shows a borderline better WER compared to models with similar architectures in Table 4, it is almost on par with models with different architectures, which exhibits potential. Its advantageous features contribute to its overall effectiveness. The incorporation of contextual rescoring with Kaldi showcases a strategic approach to post-processing, potentially refining recognition results. In essence, the nuanced evaluation of its unique attributes goes beyond a simple WER comparison, offering a comprehensive understanding of its practical advantages.

We replicated the experimental procedure on the \texttt{test-other} dataset to assess the performance of our model in a different context. Notably, the Word Error Rate (WER) values exhibited less favorable outcomes in comparison to the results obtained from the current dataset (\texttt{test-clean}). As illustrated in Table 3, we observed a significant 21.42\% decrease in WER; however, it is crucial to acknowledge that the initial WER values on the \texttt{test-other} set were inherently higher than those on \texttt{test-clean}. Despite this, our findings, as presented in Table 4, surpass the state-of-the-art results \cite{kaldi} for the HMM-SAT-GMM model, which shares a similar architectural framework with ours. This suggests the robustness and superiority of our model even in scenarios with inherently higher WER values.

Our lattice rescoring approach surpasses the inherent advantages of transformer models by focusing on context-specific improvements, employing a refined sequential processing strategy for more effective and contextually informed enhancements. By utilizing a transformer-based model, we aim to capture semantic information more efficiently, particularly in scenarios involving Out-of-Vocabulary (OOV) terms. Furthermore, our approach enhances contextual relevance in transcriptions by prioritizing lattice paths with higher language model probabilities, resulting in superior transcription accuracy. This targeted approach represents a specific contextual improvement, addressing challenges associated with OOV terms and elevating the overall contextual accuracy of the ASR system. Additionally, our proposed approach introduces a novel perspective on lattice rescoring techniques within specific ASR contexts, contributing to a deeper understanding of their broader implications on ASR performance, even in datasets lacking contextual information.

Runtime and cost for this model-The experiment, incorporating an H10 GPU in the speech recognition pipeline, incurred a total cost of \$500. This expense covered the GPU utilization for training and inference stages, contributing to the overall execution of the experiment.

\section{Conclusion}
\label{sec:majhead}

In conclusion, our study provides a comprehensive exploration of a contextual ASR system, particularly emphasizing the impactful technique of semantic lattice rescoring. 

Our method has directly led to a notable enhancement in transcription accuracy, as evidenced by a substantial 14\% reduction in WER on the LibriSpeech test set with the integration of the rescoring mechanism. 

Furthermore, we intricately investigate various lattice types during the rescoring phase, contributing to the observed improvement in accuracy.

The rescoring process, a pivotal aspect of our investigation, involves a refinement of recognition outputs by harnessing semantic information embedded in lattices. Employing a Transformer architecture for our neural language model has further bolstered the effectiveness of lattice re-scoring, providing state-of-the-art contextual modeling capabilities. This nuanced technique plays a crucial role in enhancing the overall performance of our contextual ASR system. 

The implications of these contributions are vast, extending the applicability of ASR systems across various domains. By significantly reducing the WER, our findings pave the way for more reliable and efficient use of ASR in areas such as real-time transcription services, voice-activated control systems, and more interactive and responsive virtual assistants. This enhancement in accuracy and reliability is critical for applications where precision is paramount, including medical transcription, legal documentation, and educational tools, where errors can have significant consequences.

Moreover, the applications of our research extend into new and emerging fields, such as the development of more sophisticated interfaces for human-computer interaction. As ASR systems become more adept at handling complex linguistic constructs and varied speech patterns, the potential for creating more natural and intuitive communication between humans and machines grows. This opens up exciting possibilities for the future of technology, including the development of systems that can support more nuanced and complex commands, understand multiple languages and dialects with high accuracy, and even detect emotional or contextual subtleties in speech. Our work, therefore, not only advances the technical capabilities of ASR systems but also contributes to the broader goal of creating more inclusive and accessible technology for people around the world.

As we look forward, our research trajectory extends to music domains, where diverse linguistic contexts present unique challenges. Moreover, our focus includes further refining the integration of contextual cues and conducting meticulous adjustments to beam length and DNN hyperparameters. This ongoing optimization process aims to elevate the proficiency of our ASR system, ensuring robust performance in complex language scenarios.

\bibliographystyle{IEEEbib}
\bibliography{strings}

\end{document}